\documentclass[11pt]{article}

\usepackage[final]{acl}

\usepackage{times}
\usepackage{latexsym}

\usepackage[T1]{fontenc}

\usepackage[utf8]{inputenc}

\usepackage{microtype}

\usepackage{inconsolata}

\usepackage{graphicx}

\usepackage{multirow}
\usepackage[table]{xcolor}
\usepackage{tabularx}
\newcommand{\code}[1]{\texttt{#1}}

\title{Constructing Synthetic Instruction Datasets for Improving Reasoning in Domain-Specific LLMs: A Case Study in the Japanese Financial Domain}

\author{
 \textbf{Yuma Okochi\textsuperscript{1}},
 \textbf{Fabio Milentiansen Sim\textsuperscript{2}},
 \textbf{Tomoyasu Okada\textsuperscript{1}}
\\
\\
 \textsuperscript{1}Nomura Research Institute, Ltd.,
 \textsuperscript{2}NRI Indonesia
\\
 \small{
   \textbf{Correspondence:} \href{mailto:t3-okada@nri.co.jp}{t3-okada@nri.co.jp}
 }
}

\begin{document}
\maketitle
\begin{abstract}
In adapting LLMs to specific domains, achieving both domain expertise and reasoning ability remains an urgent challenge. This study proposes a general method for constructing high-quality synthetic instruction data for any domain, starting from domain-specific vocabulary. As a demonstration, we applied this method to the financial domain and constructed a large-scale instruction dataset totaling approximately 9.5 billion tokens with Chain-of-Thought reasoning traces. Evaluation results confirmed performance improvements over baseline models on financial benchmarks, demonstrating the effectiveness of our approach. We also report findings on the impact of reasoning trace length on performance and its limitations. Lastly, we open-source our models and datasets on \href{https://huggingface.co/nri-ai}{HuggingFace}.
\end{abstract}

\section{Introduction}
\label{sec:intro}

As large language models (LLMs) are increasingly deployed in society, adaptation to domains requiring high expertise, such as finance and law, is an active research topic. Traditionally, Continued Pre-Training has been widely used for adaptation to these domains, achieving significant results in acquiring domain knowledge \citep{wuBloombergGPTLargeLanguage2023}.
In addition, recent focus has shifted beyond mere knowledge acquisition to addressing practical tasks requiring advanced logical reasoning, such as complex event analysis and future prediction (e.g., fraud detection and earnings forecasting \citep{sugiuraEDINETBenchEvaluatingLLMs2025}).

Recently, reasoning models that improve performance by generating Chain-of-Thought (CoT) reasoning traces before producing answers have attracted attention \citep{openaiOpenAIO1System2024a}. While Supervised Fine-tuning (SFT) with reasoning traces is effective for acquiring this capability, and general-purpose datasets exist \citep{NemotronPostTrainingDatasetV1,bercovichLlamaNemotronEfficientReasoning2025}, methods for constructing large-scale data that are both domain-specific (particularly in Japanese) and include reasoning traces have not been well established.

Therefore, this study proposes a data construction pipeline aimed at acquiring reasoning capabilities in specific domains, making the following contributions:

\begin{itemize}
    \item We established a method for synthesizing and filtering instruction data with reasoning traces using Reasoning LLMs, starting from domain-specific topic words. This method is applicable to diverse domains beyond finance.
    \item We applied the proposed method to the Japanese financial domain and constructed a dataset totaling approximately 9.5 billion tokens. Training on this dataset achieved performance improvements surpassing official instruction-tuned models on financial benchmarks.
    \item We analyzed the effect of controlled reasoning trace length on performance in domain-specific tasks and discuss its limitations.
\end{itemize}

\begin{figure*}[t]
  \centering
  \includegraphics[width=0.96\linewidth]{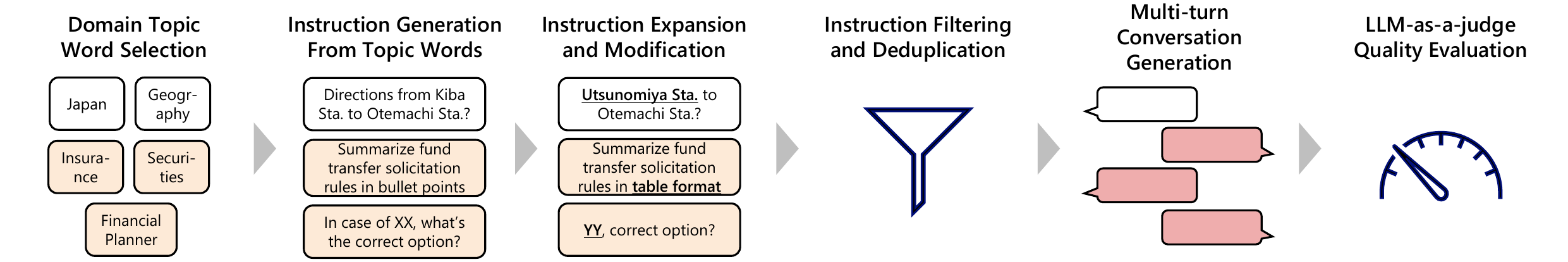}
  \caption{Proposed dataset construction pipeline}
  \label{fig:method-pipeline}
\end{figure*}

\section{Related Work}
\label{sec:related}

\subsection{Domain Adaptation and Instruction Datasets}

Instruction tuning with domain-specific datasets has been actively researched as an approach for adapting LLMs to specific domains. In the financial domain, \citet{ke-etal-2025-demystifying} and \citet{lee-etal-2024-finale} have proposed large-scale instruction datasets, though these primarily focus on English and Chinese.
For Japanese, \citet{tanabe-jafin-2024} proposed a finance-specific instruction dataset, but it mainly uses formats without reasoning traces (CoT).
This study takes an approach to solve the gap between ``domain knowledge'' and ``reasoning traces'' that was lacking in prior research, using synthetic data generation techniques. A distinctive feature of our work is presenting an extensible framework applicable to any domain by adopting a topic word-based generation method.

\subsection{Validation of Reasoning Trace Length}

For complex tasks, methods that generate reasoning traces before outputting answers are effective, and many reasoning models that output reasoning traces have been proposed \citep{openaiOpenAIO1System2024a, deepseek-aiDeepSeekV3TechnicalReport2025a, guoDeepSeekR1IncentivizesReasoning2025,yangQwen3TechnicalReport2025a}. Prior research on mathematical benchmarks has reported performance improvements with increased reasoning trace length \citep{muennighoff-etal-2025-s1}, while it has also been noted that performance may decrease beyond a certain length \citet{ghosalDoesThinkingMore2025}.
These have been primarily limited to mathematical tasks, and validation on domain-specific tasks requiring specialized knowledge and regulatory understanding has not been sufficiently conducted. This study verifies the relationship between reasoning trace length and performance using financial benchmarks.

\section{Dataset Construction Method}
\label{sec:method}

The dataset construction pipeline proposed in this study is shown in Figure~\ref{fig:method-pipeline}. This method extends the synthetic approach used by Nemotron-4 340B \citep{nvidiaNemotron4340BTechnical2024} and optimizes it for efficiently learning domain-specific knowledge and reasoning capabilities.
While this method is a domain-agnostic general process, we set the financial domain as the target for this experiment to verify its effectiveness. Unless otherwise noted, OpenAI's gpt-oss-120b \citep{openaiGptoss120bGptoss20bModel2025a} was used as the LLM for synthesis.

Additionally, detailed parameter settings for generation counts, filtering thresholds, and other specifications for each step in this pipeline are summarized in Appendix~\ref{app:dataset-pipeline-parameters}.

\subsection{Domain Topic Word Selection}
The first stage of this method is selecting topic words that comprehensively cover the target domain. We selected seed words such as industry sectors, financial products, and related technologies in the financial domain.
Additionally, to mitigate degradation of general capabilities (Catastrophic Forgetting) due to domain specialization, we mixed seed words from domains other than the target and generated sub-topic words related to each word.

\subsection{Instruction Generation From Topic Words}
Next, we had the LLM generate user prompts (instructions) related to each generated sub-topic word. To ensure task diversity, we defined four types: open-ended questions, mathematical reasoning (calculation problems), creative writing, and multiple-choice questions. For each type, we generated a sufficient number of instructions to ensure diversity.

\subsection{Instruction Expansion and Modification}
To expand the variation of user prompts, we performed expansion and modification on the instructions generated in the previous stage. Specifically, we added context, converted instructions into different formats and styles, elaborated them for specific cases, and rewrote them to target related topics. We allowed the LLM to choose the modification type and generated variants in proportion to the number of instructions in each task type.

\subsection{Instruction Filtering and Deduplication}
To ensure the quality of generated data, we performed filtering based on n-grams and word count, and deduplication using MinHash and LSH (Fuzzy deduplication), following the method of Swallow Corpus v2 \citep{hattori-2025-swallow-v2}. For n-gram filtering, we used the same parameters as prior research, and for word count filtering, we performed word segmentation using MeCab \citep{kudo-etal-2004-applying} and removed extremely short instructions.

\subsection{Multi-turn Conversation Generation}
Using the filtered instructions obtained in the previous stage, we generated multi-turn dialogues to improve interactive task reasoning capabilities. For the first turn, the instruction was input as-is, and for subsequent turns, we provided the previous exchanges and alternately generated user prompts and LLM responses including reasoning traces.

\subsection{Quality Filtering via LLM-as-a-Judge}
Finally, to determine whether the generated data accurately reflects domain knowledge and contains logical reasoning, we performed quality filtering using LLM-as-a-Judge. gpt-oss-120b was used for judgment, and low-quality data was excluded based on the average of scores from multiple evaluation criteria. Details of the evaluation method and threshold settings are shown in Appendix~\ref{app:llm-judge}.

\subsection{Utilization of Public Datasets}
In addition to domain-specific capabilities, we utilized publicly available instruction datasets to reinforce fundamental mathematical and instruction-following abilities. For math, we used NuminaMath-CoT \citep{numina_math_datasets} and Nemotron-Post-Training-Dataset-v1 \citep{NemotronPostTrainingDatasetV1,bercovichLlamaNemotronEfficientReasoning2025}, and for instruction-following, we used smol-constraints \citep{allalSmolLM2WhenSmol2025}.
We extracted only the question portions from these datasets and applied the processing from filtering/deduplication onward as shown in Figure~\ref{fig:method-pipeline}. Note that these were limited to single-turn conversations.

The final dataset consists of approximately 1.44 million samples totaling approximately 9.5 billion tokens. Dataset details are shown in Appendix~\ref{app:dataset-detail}.

\section{Training and Evaluation}
\label{sec:exp}

We trained and evaluated LLMs using the constructed dataset to verify the effectiveness of the proposed method in the target domain.

\subsection{Continued Pre-Training on Financial Corpora}
\label{sec:cpt}
In LLM domain adaptation, it is common to perform Continued Pre-Training (CPT) on pre-trained models using domain-specific corpora \citep{shiContinualLearningLarge2025}. In this study, we applied financial domain classification and quality filtering to public corpora such as Common Crawl and constructed a Japanese financial corpus. Using this corpus, we performed continued pre-training on Qwen3-14B-Base \citep{yangQwen3TechnicalReport2025a} and gpt-oss-20b \citep{openaiGptoss120bGptoss20bModel2025a}.

\subsection{Instruction Tuning via SFT}
\label{sec:sft}
To verify the effectiveness of the dataset constructed in this study, we performed Supervised Fine-tuning (SFT) on both the adopted base models and the continued pre-trained models constructed in Section~\ref{sec:cpt}.
The number of epochs was set to 2, and training was conducted on one AWS p5en.48xlarge instance (NVIDIA H200 Tensor Core GPU x 8). Each model required approximately 240 hours of training time.

\subsection{Evaluation on Financial Benchmarks}
\label{sec:fin-bench-eval}
We evaluated the trained models' capabilities in the financial domain using japanese-lm-fin-harness \citep{hirano-2024-construction} and pfmt-bench-fin-ja \citep{Hirano-pfmt}.
japanese-lm-fin-harness is a benchmark including sentiment analysis of securities reports and financial certification exam questions, which we used to verify the degree of financial knowledge acquisition.

pfmt-bench-fin-ja is a benchmark that compares whether models can appropriately respond to user queries across multiple turns for tasks such as composition, information extraction, and idea generation in the financial domain, which we used to verify financial dialogue capabilities.

The evaluation methods for each benchmark were adapted from the public repository implementations, with details shown in Appendix~\ref{app:fin-bench-eval-detail}.

\begin{table*}[t]
    \centering
    \fontsize{8pt}{10pt}\selectfont
    \begin{tabular}{ll|cccccc|ccc}
    \hline
    \multirow{2}{*}{Model} & \multirow{2}{*}{CoT} &
    \multicolumn{6}{c|}{\textbf{japanese-lm-fin-harness}} &
    \multicolumn{3}{c}{\textbf{pfmt-bench-fin-ja}} \\
    \cline{3-8}\cline{9-11}
     &  & Avg. & chabsa & cma & cpa & fp2 & ss1 & Avg. & turn1 & turn2 \\
    \hline
    Qwen3-14B & No &
    \cellcolor{gray!15}66.50 & 91.73 & 87.58 & 41.83 & 47.03 & 64.31 &
    \cellcolor{gray!15}7.781 & 7.819 & 7.743 \\
    Qwen3-14B & Yes &
    \cellcolor{gray!15}71.04 & \textbf{91.96} & \textbf{93.26} & \textbf{49.37} & 53.37 & 67.22 &
    \cellcolor{gray!15}8.104 & 8.211 & 7.997 \\
    \hline
    Qwen3-14B-Base + SFT (Ours) & Yes &
    \cellcolor{gray!15}70.69 & 91.48 & 91.20 & 47.96 & 55.00 & \textbf{67.81} &
    \cellcolor{gray!15}8.415 & 8.472 & 8.358 \\
    Qwen3-14B-Base + CPT + SFT (Ours) & Yes &
    \cellcolor{gray!15}\textbf{71.78} & 91.62 & 91.45 & 48.59 & \textbf{60.00} & 67.27 &
    \cellcolor{gray!15}\textbf{8.455} & \textbf{8.514} & \textbf{8.395} \\
    \hline \hline
    gpt-oss-20b & No &
    \cellcolor{gray!15}61.17 & 91.09 & 81.33 & 33.45 & 41.82 & 58.17 &
    \cellcolor{gray!15}7.480 & 7.425 & 7.534 \\
    gpt-oss-20b & Yes &
    \cellcolor{gray!15}66.93 & 91.80 & 90.46 & 38.51 & 49.74 & 64.15 &
    \cellcolor{gray!15}7.883 & 7.858 & 7.908 \\
    \hline
    gpt-oss-20b + SFT (Ours) & Yes &
    \cellcolor{gray!15}69.56 & 91.87 & 90.87 & 43.65 & 52.63 & \textbf{68.80} &
    \cellcolor{gray!15}\textbf{8.432} & \textbf{8.439} & \textbf{8.424} \\
    gpt-oss-20b + CPT + SFT (Ours) & Yes &
    \cellcolor{gray!15}\textbf{72.50} & \textbf{91.89} & \textbf{94.24} & \textbf{45.51} & \textbf{62.71} & 68.15 &
    \cellcolor{gray!15}8.209 & 7.992 & 8.425 \\
    \hline
    \end{tabular}
    \caption{Evaluation results of each model on financial benchmarks (CoT = presence of reasoning traces)}
    \label{tab:fin-benchmark}
\end{table*}

The evaluation results for each benchmark are shown in Table~\ref{tab:fin-benchmark}.
When only SFT was performed, the Qwen3-14B-Base model showed lower scores compared to the official instruction-tuned model. Deeper analysis revealed that the lower scores were largely attributable to answer-extraction failures, such as repetitive outputs or non-compliance with the specified answer format. However, during the training process leading to the final results, a trend was observed where these errors decreased as the number of training steps increased, so performance improvement is expected with continued training.

When applying SFT on our dataset to the CPT models, the average scores improved on both benchmarks. gpt-oss-20b exceeded the official instruction-tuned model on all subtasks, and Qwen3-14B showed similar trends except for some tasks. This suggests that the proposed data construction method can effectively connect domain knowledge injection (CPT) with reasoning capability acquisition (SFT) to maximize task performance in specific domains.

We also compared performance with and without reasoning for the baseline models. We found that inference with reasoning traces improved performance by approximately 4.5–5.7 points on japanese-lm-fin-harness and by approximately 0.4 points on pfmt-bench-fin-ja across all models, confirming the effectiveness of reasoning-enabled inference on financial domain benchmarks.

\subsection{Validation of Performance Improvement with Reasoning Trace Length}
\label{sec:fin-test-time-scaling-eval}

From the evaluation results in Section~\ref{sec:fin-bench-eval}, it was confirmed that outputting reasoning traces contributes to performance improvement on financial tasks. To verify the impact of reasoning trace length on model performance, we fixed the reasoning trace length to certain values and varied these fixed values in stages.

Following prior research \citep{muennighoff-etal-2025-s1}, we used the following procedure to fix the reasoning trace length to specified values:

\begin{description}
    \item[When ending shorter than specified length] Remove the reasoning end token (e.g., \code{</think>}) and add \code{``Wait,''} to the end to continue inference.
    \item[When ending at or beyond specified length] Delete tokens exceeding the specified length, insert the reasoning end token and text prompting the final answer (newline) \code{``Final Answer:''}, and then generate the final answer anew.
\end{description}

\begin{figure}[t]
    \centering
    \includegraphics[width=0.96\linewidth]{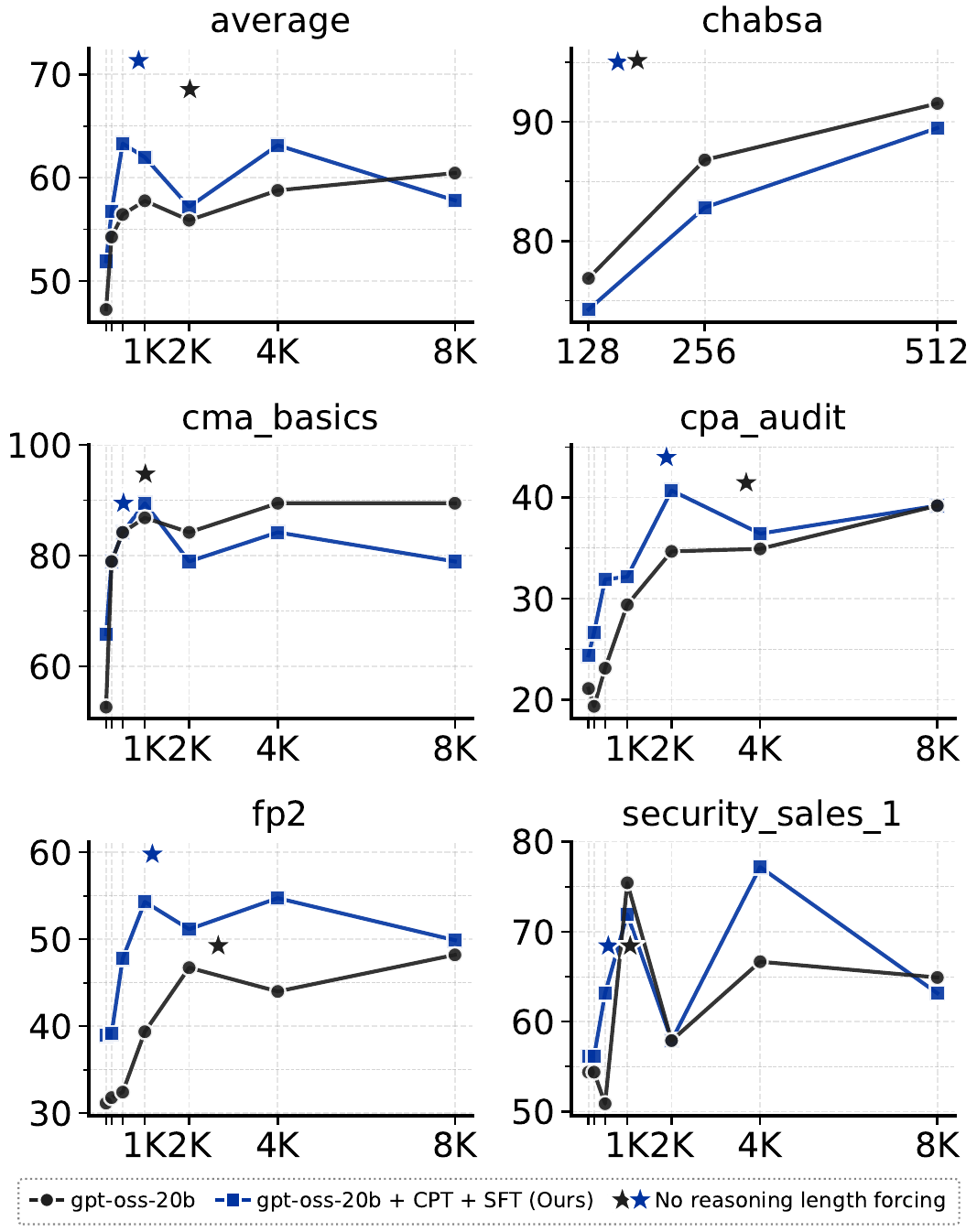}
    \caption{Changes in task accuracy with reasoning trace length}
    \label{fig:reasoning-scale}
\end{figure}

The target benchmark is japanese-lm-fin-harness. For the four subtasks other than chabsa, we set fixed lengths at 7 levels: 128, 256, 512, 1024, 2048, 4096, and 8192 tokens. For chabsa, due to execution time constraints, we limited experiments to 3 levels: 128, 256, and 512 tokens.

The execution results for each task are shown in Figure~\ref{fig:reasoning-scale}.
For each subtask except chabsa, performance improved by extending the fixed length up to 1024 tokens. For chabsa as well, a similar trend was observed up to 512 tokens within the verification range. However, no significant performance improvement was confirmed when setting the fixed length to 2048 tokens or more (diminishing returns).
Analysis of reasoning traces revealed the following behaviors as factors:

\begin{description}
    \item[Early termination of reasoning] Even when inserting \code{``Wait,''}, the model transitions to a conclusion immediately after, such as \code{``Wait, \textbf{but} the answer is D. So the final answer is D.''}, attempting to end the reasoning trace without deeper reasoning.
    \item[Output loops] Especially at 4096 tokens or more, output repeating only confirmation of conclusions continues until reaching the fixed length, resulting in no substantial reasoning being performed.
\end{description}

Additionally, when forcibly terminating reasoning traces, a performance degradation of several points was confirmed across all tasks compared to inference without forced termination (marked with $\star$ in Figure~\ref{fig:reasoning-scale}). The fact that performance degrades even with approximately the same reasoning trace length suggests that the method of terminating reasoning traces itself may affect performance. For example, methods using natural termination phrases as verified with Qwen3 \citep{yangQwen3TechnicalReport2025a}, such as
\code{``Considering the limited time by the user, I have to give the solution based on the thinking directly now\textbackslash n</think>\textbackslash n\textbackslash n''}
could be considered, but examination of various countermeasures including other termination methods is needed in the future.

\section{Conclusion}
\label{sec:conclusion}

This study proposes a general framework for constructing synthetic instruction datasets aimed at improving LLM reasoning capabilities in specific domains.
To demonstrate the proposed method, we constructed a dataset totaling approximately 9.5 billion tokens with reasoning traces targeting the Japanese financial domain, and evaluation results achieved performance surpassing official instruction-tuned models. This shows that synthetic data generation starting from topic words is extremely effective for domain adaptation.
Additionally, analysis of reasoning trace length in domain-specific tasks confirmed a trend where performance improves up to around 1024 tokens but plateaus thereafter.
This method is easily extensible to other domains with high expertise beyond finance, and we hope that this approach can serve as a foundation for future work on building domain-specific LLMs.

\section*{Acknowledgments}

This work was supported by the Ministry of Economy, Trade and Industry (METI) and the New Energy and Industrial Technology Development Organization (NEDO) under the ``Research and Development Project of the Enhanced Infrastructures for Post-5G Information and Communication Systems'' through the Generative AI Accelerator Challenge (GENIAC).

We would also like to express our gratitude to Assistant Professor Tatsunori Hashimoto (Stanford University) for his valuable advice in shaping the research concept.

\bibliography{custom}

\appendix

\section{Dataset Pipeline Parameter Details}
\label{app:dataset-pipeline-parameters}

The detailed configuration parameters for each step in this study's dataset construction pipeline are shown in Table~\ref{tab:pipeline-params}.

\begin{table*}[h]
    \centering
    \small
    \begin{tabular}{l|l|r}
    \hline
    \textbf{Step} & \textbf{Parameter} & \textbf{Value} \\
    \hline
    \multirow{2}{*}{1. Topic word selection} & Financial domain seed words & 135 \\
    & General domain (for mixing) seed words & 20 \\
    \hline
    \multirow{2}{*}{2. User question generation} & Generation count / sub-topic (multiple-choice) & 8 \\
    & Generation count / sub-topic (others) & 10 \\
    \hline
    \multirow{2}{*}{3. Instruction expansion} & Expansion variations / question (multiple-choice) & 3 \\
    & Expansion variations / question (others) & 5 \\
    \hline
    4. Filtering & Minimum word count (exclude below this) & 10 \\
    \hline
    5. Multi-turn generation & Maximum turns & 3 \\
    \hline
    \end{tabular}
    \caption{Dataset construction pipeline configuration parameters}
    \label{tab:pipeline-params}
\end{table*}

\section{LLM-as-a-Judge Quality Filtering Details}
\label{app:llm-judge}

We performed quality evaluation and filtering using LLM-as-a-Judge on the generated multi-turn interactive instruction dataset. gpt-oss-120b was used as the judge model, and we evaluated whether the LLM responses in each sample were appropriate from the five evaluation perspectives shown in Table~\ref{table:quality-dimensions}. We performed 5-level evaluation (1--5 points) for each perspective, instructing the evaluator model to use objective and strict criteria.

\begin{table*}[h]
    \centering
    \small
    \begin{tabularx}{\textwidth}{l|X}
    \hline
    \textbf{Perspective} & \textbf{Criteria (5 levels)} \\
    \hline
    Accuracy (factual correctness, presence of misinformation) & 5: Completely accurate / 3: Some errors / 1: Major misinformation \\
    Relevance/Instruction-following in response & 5: Completely appropriate / 3: Partially unaddressed / 1: Off-topic \\
    Usefulness/Comprehensiveness of answer/information & 5: Extremely useful / 3: Average / 1: Insufficient \\
    Reasoning/Depth quality, analysis, logical consistency & 5: Excellent reasoning / 3: Average / 1: Logical breakdown \\
    Safety/Appropriateness, ethics of expression & 5: Completely safe / 3: Minor issues / 1: Dangerous \\
    \hline
    \end{tabularx}
    \caption{Quality evaluation perspectives and criteria}
    \label{table:quality-dimensions}
\end{table*}

The evaluation results on the dataset showed that the percentages achieving 5 points for each perspective were: Accuracy 81.7\%, Relevance/Instruction-following 97.3\%, Usefulness/Comprehensiveness 97.6\%, Reasoning/Depth 91.2\%, and Safety/Appropriateness 99.95\%. Accuracy was the lowest, confirming the occurrence of factual errors and hallucinations. In this study, to ensure maximum quality of SFT data, we set strict filtering criteria to select only samples that achieved 5 points on all 5 perspectives. This criterion selected 632,636 samples (75.4\%) from 839,398 samples, with approximately 24.6\% excluded. Excluded samples contained issues such as repetition (repeating the same phrases), inappropriate responses to instructions, and breakdown of reasoning processes.

\section{Dataset Details}
\label{app:dataset-detail}

Information on sample counts, token counts, and presence of multi-turn for the datasets used in training is shown in Table~\ref{table:dataset-details}.

\begin{table*}[h]
    \centering
    \fontsize{7pt}{9pt}\selectfont
    \setlength{\tabcolsep}{3pt}
    \begin{tabular}{l c r | r | r r r r | c}
    \hline
    \multirow{2}{*}{\textbf{Dataset Name}} & \multirow{2}{*}{\textbf{Category}} & \multirow{2}{*}{\textbf{Samples}} & \multirow{2}{*}{\textbf{Total Tokens}} &
    \multicolumn{4}{c|}{\textbf{Avg. Tokens (/sample)}} &
    \multirow{2}{*}{\textbf{Multi-turn}} \\
    \cline{5-8}
     &  &  &  &
    user & assistant & reasoning & total & \\
    \hline
    nri-fin-reasoning & Finance/General & 632,636 & 6,352,341,377 & 191 & 8,996 & 855 & 10,041 & Yes \\
    NuminaMath-CoT-modified & Math & 88,727 & 141,678,621 & 83 & 508 & 1,006 & 1,596 & No \\
    Nemotron-Post-Training-math-modified & Math & 694,168 & 2,999,298,386 & 169 & 640 & 3,512 & 4,320 & No \\
    smol-constraints-modified & Instruction-following & 24,411 & 12,803,102 & 71 & 199 & 254 & 524 & No \\
    \hline
    \textbf{Total} & --- & \textbf{1,439,942} & \textbf{9,506,121,486} & \textbf{171.7} & \textbf{4,295.4} & \textbf{2,134.6} & \textbf{6,601.7} & --- \\
    \hline
    \end{tabular}
    \caption{Statistics of each dataset used in training}
    \label{table:dataset-details}
\end{table*}

\section{Modifications to Financial Benchmark Evaluation Methods}
\label{app:fin-bench-eval-detail}

For japanese-lm-fin-harness, the public repository implementation uses log-likelihood of tokens corresponding to choices for judgment, but this makes evaluation unstable for instruction models that output reasoning traces, so we modified the prompts to chat format and had models answer within \code{\textbackslash boxed\{\}}.
Additionally, since recent models tend to specify particular temperature and top-p sampling parameters to stabilize deep reasoning, we did not use deterministic decoding via greedy method but instead performed evaluation in pass@1 format, averaging over multiple attempts.

For pfmt-bench-fin-ja, unlike the public repository implementation, we used GPT-5 mini (version: 2025-08-07) as the model for LLM-as-a-judge, which has higher performance.

\end{document}